\pdfoutput=1

\documentclass[11pt,a4paper]{article}
\usepackage[hyperref]{acl2021}
\usepackage{breakurl}
\usepackage{times}
\usepackage{latexsym}

\usepackage{microtype}

\usepackage{tabularx}
\usepackage{booktabs}
\usepackage{amsmath}
\usepackage{listings}
\usepackage{newfloat}
\usepackage[framemethod=TikZ]{mdframed}
\usepackage{multirow}
\usepackage{enumitem}

\usepackage{pifont}
\newcommand{\cmark}{\ding{51}}%
\newcommand{\xmark}{\ding{55}}%

\usepackage{xcolor}
\usepackage[normalem]{ulem}

\def\ODdel#1{\bgroup\markoverwith{\textcolor{cyan!80!yellow!80!black!100}{\rule[0.4ex]{2pt}{3pt}}}\ULon{#1}}

\aclfinalcopy 

\title{Shades of BLEU, Flavours of Success: The Case of MultiWOZ}

\author{Tomáš Nekvinda \normalfont{and} \textbf{Ondřej Dušek} \\
  Charles University, Faculty of Mathematics and Physics \\
  Institute of Formal and Applied Linguistics \\
  Prague, Czech Republic \\
  \texttt{\{nekvinda,odusek\}@ufal.mff.cuni.cz} \\}

\date{}

\begin{document}
\maketitle

\begin{abstract}
The MultiWOZ dataset \cite{mwz20} is frequently used for benchmarking context-to-response abilities of task-oriented dialogue systems. In this work, we identify inconsistencies in data preprocessing and reporting of three corpus-based metrics used on this dataset, i.e., BLEU score and Inform \& Success rates. We point out a few problems of the MultiWOZ benchmark such as unsatisfactory preprocessing, insufficient or under-specified evaluation metrics, or rigid database. We re-evaluate 7 end-to-end and 6 policy optimization models in as-fair-as-possible setups, and we show that their reported scores cannot be directly compared. To facilitate comparison of future systems, we release our stand-alone standardized evaluation scripts. 
We also give basic recommendations for corpus-based benchmarking in future works. 
\end{abstract}

\section{Introduction}

While human judgements are irreplaceable in dialogue systems evaluation and using full dialogue evaluation instead of evaluating isolated responses given ground-truth contexts cannot fully measure system performance \cite{liu16,takanobu_is_2020},
corpus-based evaluation metrics, such as BLEU and corpus-based entity match and success rate \cite{wen17}, are still very important for model development and are often used to compare models and establish state-of-the-art. We show on the MultiWOZ benchmark \cite{mwz20}, one of the most frequently used and most challenging dialogue system datasets today, that these comparisons do not hold if several basic conditions are not met, and that these conditions are not met for most of the recent works using corpus-based evaluation on this dataset. This means the assessment of progress in terms of dialogue modeling is obscured by noise coming from differences in preprocessing or metrics implementation variants. 

This paper is not a critique of the MultiWOZ benchmark or of systems evaluated on it. Instead, it is a call for consistency and increased rigor in automatic evaluation. In addition to providing the analysis and identifying problems with the benchmark and current state-of-the-art reporting, we include recommendations for consistency in corpus-based score comparisons. 
In particular, we advocate for: 
(1) using standardized implementations of metrics;
(2) evaluating either on detokenized surface texts, or using standardized preprocessing and postprocessing;
(3) reporting the exact scripts used for evaluation;
(4) release of system outputs.
We also show that there is room for additional metrics of output diversity, and we
add an observation on the overlap between the dialogue goals and states in training and test sections of the MultiWOZ data. 

Our work can be summarized as follows:

\begin{itemize}[itemsep=0pt,topsep=4pt,leftmargin=12pt]
\item We identify, list, and discuss consistency issues associated with the MultiWOZ benchmark;
\item We compare and re-evaluate 13 end-to-end or policy optimization systems, using a single implementation of metrics and preprocessing;
\item We release the outputs of all compared systems in a unified format and provide stand-alone standardized evaluation scripts that allow for consistent comparison of future works on this dataset;\footnote{\url{https://github.com/Tomiinek/MultiWOZ_Evaluation}} 
\item In addition to standard MultiWOZ corpus-based metrics, we evaluate all systems in terms of the diversity of their outputs.
\end{itemize}

\section{Related Work}
\label{sec:related}

Most works on evaluation methods in dialogue response generation \cite{deriu_survey_2020} focus on human evaluation \cite{walker_paradise:_1997}, e.g., choosing the best methodology with respect to quality and consistency \cite{santhanam_towards_2019} or robustness \cite{dinan_build_2019}. Recent surveys in natural language generation reflect on divergence and inconsistency in human evaluation practice 
\cite{howcroft_twenty_2020,belz_disentangling_2020}, in a similar spirit to our examination, but on a broader scale.

Despite the availability of simulator evaluation \cite{schatzmann_survey_2006, young_hidden_2010, convlab}, corpus-based metrics have been the go-to evaluation method in end-to-end neural dialogue systems since the first implementations \cite{wen17,eric_copy-augmented_2017} and are a de-facto standard until today (cf.~Section~\ref{sec:systems}). 
There are works showing problems of corpus-based metrics: limited correlation with human judgements \cite{novikova_why_2017,takanobu_is_2020} and mixed performance depending on human reference texts used \cite{freitag_bleu_2020} or evaluated systems \cite{mathur_tangled_2020}. 
Many works aim at creating more reliable metrics
 \cite{galley_deltableu:_2015}. Recent focus is
on trained neural metrics \cite{dziri_evaluating_2019,mehri_usr_2020}, but they are not yet in wide use.

Our work is probably the closest to \citet{post18}'s assessment of inconsistencies in different implementations of 
BLEU \cite[see Section~\ref{sec:mwoz-metrics}]{bleu},
calling for comparability and proposing a standard implementation.
To our knowledge, we are the first to evaluate the use of corpus-based metrics in dialogue systems in this fashion.


\section{Benchmark and Systems}
\label{sec:description}

\begin{table}[bt!]
    \centering\small
    \setlength{\tabcolsep}{4pt}
    \renewcommand{\arraystretch}{1.45}
    \begin{tabular}{p{0.075\linewidth}|p{0.85\linewidth}}
      \toprule
      \textbf{Side} & \textbf{Utterance} \\
      \midrule
      User & Can you help me find a train that will arrive in birmingham new street by 16:15? \\
      Sys. & What day would you like to travel? \\
      User & I would like to leave Friday from Cambridge. \\
      Sys. & TR2716 will leave cambridge at 13:01 and arrive at birmingham at 15:44. The trip will take 163 minutes and cost 75.10 pounds. Would you like to book? \\
      User & Yes, I would like to book for 3 people. \\
      Sys. & I booked the train. It was 225.3 and your reference is AS5SBC8P . Do you need help with anything else? \\
      User & Yes I also need information on queens' college. \\
      Sys. & It's located in west Cambridge on silver street. The postcode is cb39et. The phone number is 01223335511. The entrance fee is 2.50 pounds. Do you need anything else? \\
      User & No, that's great. Thank you. Good-bye. \\
      Sys. & Thank you for using the Cambridge TownInfo centre. \\
      \bottomrule
    \end{tabular}
    \caption{Example multi-domain conversation from the MultiWOZ dataset (conversation ID: PMUL1266).}
    \label{tab:conversation}
\end{table}

\subsection{The MultiWOZ dataset}
\label{sec:dataset}

The MultiWOZ~2.0 dataset 
\citep{mwz20}
includes about 10k task-oriented dialogues in 7 domains (restaurants, hotels, tourist attractions, trains, taxi, hospital, police) with dialogue state and system action annotation. Larger domains (restaurants, hotels, attractions, trains) have an associated database. 
The data was collected via human-to-human interaction on a crowdsourcing platform using the Wizard-Of-Oz approach \citep{wen17}. 
Crowd workers were instructed with goals such as booking or finding information about a restaurant or train (see Table~\ref{tab:conversation}). 
The dataset authors provided supporting code\footnote{
\url{https://github.com/budzianowski/multiwoz/}} and baselines for dialogue state tracking (DST), context-to-text (CTR), and action-to-text generation tasks. 

\paragraph{MultiWOZ 2.1:} 
\citet{mwz21} released an update with re-annotated dialogue states and added explicit system action annotation.

\paragraph{MultiWOZ 2.2} \cite{mwz22} has more fixes for state annotation in 
17.3\% of turns, a redefined ontology, and canonical forms for slot values (e.g. “13:00” for “1pm”) for better DST evaluation. Additionally, 
it introduces slot span annotations allowing easy delexicalization, which was previously based only on string matching heuristics. 



\subsection{Corpus-based Metrics on MultiWOZ}
\label{sec:mwoz-metrics}


All standard CTR metrics on MultiWOZ -- BLEU, Inform \& Success rate -- are 
calculated on \emph{delexicalized} texts, i.e., texts where dialogue slot values, such as venue names, are replaced by placeholders \citep{wen_semantically_2015}. While using delexicalized utterances prevents errors in venue names to affect the evaluation,
it prevents the use of an interactive human evaluation, model-based evaluation metrics known from open-domain dialogue research \citep{gao20}, or end-to-end evaluation with user simulators such as ConvLab \cite{convlab}. 
\paragraph{BLEU} 
\citep{bleu}, originally designed for machine translation (MT) evaluation, is based on comparison of n-grams in human-written references and machine-generated hypotheses. Following \citet{wen17}, BLEU is used to measure \emph{fluency} of output responses where the human utterances are used as the reference. 
Using the metric for assessing fluency of the responses is not ideal, because as opposed to the intended use of BLEU, there is only a single reference available. Moreover, the set of valid responses is arguably larger for dialogue than for MT. \citet{liu16} show that metrics adopted from MT correlate very weakly with human judgements in dialogue responses.

\paragraph{Inform \& Success rates:}  The Inform 
rate relates to \emph{informable} slots, which are attributes that allow the user to constrain database searches, e.g., restaurant location or price range. The Success rate focuses on \emph{requestable} slots, i.e., those that can be asked by the user, e.g., phone number. Both are calculated on the level of dialogues. 


\citet{su15} consider a dialogue to be successful if the evaluated system provided all of the requested information for an entity satisfying the user's constraints. Following this definition, 
\citet{wen17} set aside the Match rate describing whether the entity found at the \emph{end} of each dialogue matches the user's goal. However, MultiWOZ dialogues include multiple interleaving domains and calculating the rates only at the end is not sufficient.

Therefore, \citet{mwz20} mark a dialogue as successful if for each domain in the user's dialogue goal: (1) the last offered entity matches (satisfies the goal constraints), and (2) the system mentioned all requestable slots required by the user.  The \emph{Inform} rate then marks the proportion of dialogues complying to (1), \emph{Success} rate is the proportion of fully successful dialogues.

The offered entities and mentions of requestable slots are tracked over the delexicalized responses for the whole dialogue, making use of slot placeholders. 
If an utterance contains a slot naming an entity, e.g., restaurant name or train ID, the current dialogue state for the corresponding domain is used to query the database and an entry is sampled from the search results. At the end of a dialogue, the recorded entities and requestable slots are compared to expected values from the dialogue goal
(see Appendix~\ref{sec:appendixa} for an example).
The dialogue can thus be considered unsuccessful if the system does not mention a venue name or train ID at the right turn,\footnote{It must in practice hit the single suitable turn because responses are generated given ground-truth dialogue context.} does not track 
 the user's search constraints, or ignores the user's requests. 

\subsection{Systems Evaluating on MultiWOZ}
\label{sec:systems}

We discuss performance of 13 recent systems that use CTR evaluation on MultiWOZ -- 7 end-to-end and 6 policy-optimization systems, which use ground-truth dialogue states during training and inference. 
We include models for which we got test set predictions and systems with public code for which we managed to replicate reported results.\footnote{We were not successful in getting code, model weights, or original predictions for other systems, such as SimpleTOD \citep{simpletod}, or ARDM \citep{ardm}.}

Out of the 13 compared works, 7 only report BLEU, Inform, and Success 
with no other evaluation; 4 use human ratings of individual outputs, 
and only 2 include human evaluation on full dialogues.\footnote{Note that full interaction is not possible with policy optimization models unless an external DST model is applied.}



\begin{table*}[hbt!]
    \centering\small
    \setlength{\tabcolsep}{5.75pt}
    \renewcommand{\arraystretch}{1.2}
    \begin{tabular}{p{0.09\linewidth}|p{0.855\linewidth}}
      \toprule
      Delexical. & Utterance \\
      \midrule
      Original   & Cafe jello gallery has a free entrance fee. The address is cafe jello gallery, 13 magdalene street and the post code is cb30af. Can i help you with anything else? \\
      MWZ 2.2       & [address] has a [entrancefee] entrance fee. The address is [name], [address] and the post code is [postcode]. Can I help you with anything else?  \\
      \midrule
      HDSA          & [attraction\_name] has a free entrance fee. The address is [attraction\_address] and the post code is [attraction\_postcode]. Can i help you with anything else? \\
      DAMD          & [value\_name] has a [value\_price] entrance fee. The address is cafe jello gallery, [value\_address] and the post code is [value\_postcode]. Can i help you with anything else? \\
      AuGPT         & [address] has a free entrance fee. The address is cafe jello gallery, [address] and the post code is [postcode]. Can I help you with anything else? \\
      UniConv       & [attraction\_name] has a [attraction\_pricerange] entrance fee. The address is [attraction\_name], 13 [attraction\_address] and the post code is [attraction\_postcode]. Can i help you with anything else? \\
      LAVA          & [attraction\_name] has a free entrance fee. The address is [attraction\_name], [value\_count] [attraction\_address] and the post code is [restaurant\_postcode]. Can i help you with anything else? \\
      
      \bottomrule
    \end{tabular}
    \caption{An example utterance from the MultiWOZ dataset with different styles of delexicalization. The first row shows the non-delexicalized source response. Other styles are paired with the systems that use or introduced them.}
    \label{tab:delex}
\end{table*}

An important representative of the end-to-end systems is DAMD \citep{damd}. 
It uses a multi-action data augmentation and multiple GRU \citep{cho14} decoders. Similarly, LABES \citep{labes} employs a few GRU-based decoders, but it represents the dialog state as a latent variable. DoTS \citep{jeon2021} also uses GRUs, but the model makes use of a BERT encoder \citep{devlin19} to get a context representation. MinTL \citep{mintl} applies a diff-based approach to state updates, with backbones based on the T5 and BART models \citep{t5, bart}. UBAR is based on a fine-tuned GPT-2 model \citep{gpt2}, similarly to AuGPT \citep{augpt} which uses back-translations for response augmentation, and SOLOIST \citep{soloist} which makes use of machine teaching \citep{shukla2020}. We used author-provided outputs for SOLOIST and AuGPT, author-trained checkpoints for DoTS, LABES,\footnote{We were able to generate outputs for 91.66\% test utterances with this checkpoint. We note this in Tables~\ref{tab:bleu},~\ref{tab:rates} and~\ref{tab:diversity}.} and UBAR, and we trained DAMD and MinTL\footnote{We were only able to reproduce the T5-small model and use it in this comparison.} from scratch using publicly available code.
DAMD, MinTL and SOLOIST use MultiWOZ~2.0; the remaining models trained on the 2.1 version.
DAMD, LABES, MinTL, and UBAR are based on the same code base and use similar evaluation scripts.

We also compared 6 policy optimization models. SFN \citep{sfn}, HDNO \citep{hdno}, and LAVA \citep{lava} use reinforcement learning for training. HDSA \citep{hdsa} uses BERT and exploits the hierarchical structure of dialog acts. MarCo \citep{marco} and UniConv \cite{uniconv} generate explicit system actions in parallel with the response. We use the public predictions for LAVA and the provided pretrained models for other models. UniConv and HDNO are trained on MultiWOZ 2.1, other systems use the 2.0 version.
As opposed to end-to-end models, the version affects the evaluation because the ground-truth state is supplied to the model. 
The comparison of these systems is thus not completely fair, but we believe that the differences are small in comparison with the differences in evaluation scripts and setups (see Section~\ref{sec:results}).

\section{Benchmark Caveats}
\label{sec:drawbacks}


While MultiWOZ and the associated metrics described in Section~\ref{sec:description} represent the state-of-the-art in corpus-based dialogue evaluation practice, the benchmark has the following limitations that researchers 
need to be aware of:
(1) delexicalization problems -- imprecise delexicalization based on string matching and varying implementations thereof (Section~\ref{sec:preprocessing}), (2) lack of standardized postprocessing (i.e., lexicalization methods, Section~\ref{sec:postprocessing}), (3) database problems, i.e., multiple surface forms of database values and no information about booking availability (Section~\ref{sec:database}), (4) atypical metric implementations (Section~\ref{sec:eval-problems}), (5) lack of diversity evaluation (Section~\ref{sec:diversity}), (6) similarity between training and test data (Section~\ref{sec:folds}).



\subsection{Preprocessing} 
\label{sec:preprocessing}

CTR evaluation metrics used in the benchmark work with \emph{delexicalized} texts (see Section~\ref{sec:mwoz-metrics}). 
However, the implementation of delexicalization provided with the dataset is limited; it only applies to some expressions, leaving other slot values lexicalized.
That is why most systems use their own delexicalization methods.
The original delexicalization uses placeholders consisting of the domain name and the slot name, e.g. \emph{taxi\_phone}. Recent works following DAMD \cite{damd} remove domain names from the placeholders and determine the active domain from changes in the predicted dialogue state or model it directly.

We identified five different delexicalization styles among the 13 systems described in Section~\ref{sec:systems}. Table~\ref{tab:delex} shows a sample system turn for which the outputs of all the delexicalization approaches are different. This is a problem since all works use their own preprocessed data as references for BLEU computation.
We checked the test set for slot placeholders and found that
70.61\% of the utterances contain a slot in at least one delexicalized variant and only 17.52\% responses with slots exactly match for all the systems.\footnote{8 utterances (including the example in Table~\ref{tab:delex}) are pairwise different between all 5 delexicalizations.}

Moreover, preprocessing scripts of some works remove contracted verb forms or keep suffixes such as “-s”, “-ly” when delexicalizing nouns or adverbs, e.g., “moderately” becomes “[pricerange]-ly”.

\subsection{Postprocessing}
\label{sec:postprocessing}

The 
MultiWOZ code base does not implement backward lexicalization of texts. Out of 12 systems for which we have the source code available, only four offer 
scripts for lexicalizing slot values and thus allow further in-depth evaluation. 

 
\subsection{Database: Surface Forms and Booking}
\label{sec:database}

The original MultiWOZ implementation of the database performs only subtle normalization of the database search constraints, such as replacing “\&" with “and". However, the slot values can have multiple valid surface forms; e.g., “4pm” and “16:00” or “the botanical gardens at cambridge university" and “cambridge university botanic gardens" correspond to the same database entities. 
Database query normalization is crucial for end-to-end systems,
as opposed to the policy optimization models, which use ground-truth dialogue states with normalized values. The flexibility of the database might affect the Inform \& Success rates, because they are based on information about database entries \emph{complying} with the current dialogue state.
 
The original database does not contain any information about booking availability, because during the data collection, crowd workers were sometimes instructed to refuse a booking at a specific time, ask for another place, etc., and accept the booking with new constraints. This brings a problem into the evaluation, because some works use the ground-truth booking information (mined from the dialogue state and system action annotations) even during evaluation, whereas other ignore it and let their systems behave randomly.

\subsection{Evaluation}
\label{sec:eval-problems}

\paragraph{BLEU:} 

The original MultiWOZ BLEU implementation internally uses a trivial tokenization splitting on whitespace.
However, current models often use subword tokenization and complex detokenization to remove any redundant whitespace \citep{subwords, spiece}. This new-style detokenization might produce words with leading or trailing punctuation. Some works ignore this fact completely, or use an alternative BLEU implementation, including tokenization, from NLTK \citep{nltk}.

\begin{table*}[ht]
    \centering\small
    \setlength{\tabcolsep}{4.5pt}
    \renewcommand{\arraystretch}{0.9}
    \begin{tabular}{l|cc|cccc}
      \toprule
      \multicolumn{1}{c|}{\multirow{2}{*}{System}} & \multicolumn{2}{c|}{\multirow{1}{*}{BLEU score}} &
      \multicolumn{4}{c}{\multirow{1}{*}{Inform \& Success rate}}  \\
      & Delexical. & Tokenization  & Venue comparison & Venue updates & Reduced search & Domain source \\
      \midrule
      DAMD     & DAMD  & word     & intersection & name, id & \cmark & state change \\
      MinTL    & DAMD  & sub-word & intersection & name, id & \cmark & state change \\
      UBAR     & DAMD  & sub-word & intersection & name, id & \cmark & state change \\
      SOLOIST  & HDSA  & sub-word & - & - & - & slot names \\
      AuGPT    & AuGPT & sub-word, NLTK & first & end & \xmark & predicted \\
      LABES    & DAMD  & word     & intersection & name, id & \cmark & state change \\
      DoTS     & HDSA  & word     & sampling & name, id & \xmark & slot names \\
      \midrule
      MarCo     & HDSA & word, NLTK & subset & name, id & \xmark & slot names \\
      HDSA      & HDSA & word, NLTK & subset & name, id & \xmark & slot names \\
      HDNO      & HDSA & word & sampling & name, id  & \xmark & slot names \\
      SFN       & HDSA & word & sampling & name, id & \xmark & slot names \\
      UniConv  & UniConv & word & sampling & name, id, ref. & \xmark & slot names \\
      LAVA      & LAVA & word & sampling & name, id & \xmark & slot names \\
      \bottomrule
    \end{tabular}
    \caption{Setups of compared systems with respect to the used delexicalization method, tokenization, and Inform \& Success implementation. The “Venue comparison” column describes the method of comparing offered and goal database entries, “Venue updates” indicates when the set of database entries complying to the current state is updated, “Reduced search” reflects the database implementation that ignores other search constraints if a venue name or train ID is present, and “Domain source” describes the source of information about the active turn domain. }
    \label{tab:setups}
\end{table*}

\paragraph{Inform \& Success rate:} 
We found two main problems here. The first one comes from random database entry sampling -- if multiple entities match the dialogue state, one of them is sampled at random from the database results. The set of entries complying with the dialogue state does not have to be a subset of the ground-truth set of entries complying with a given prescribed user goal from the test set. If the database results and the ground-truth set have an imperfect overlap, the sampling may choose an entry from the difference of the two sets, which is counted as a failure. However, if an entry from the intersection of the two sets is chosen, it counts as a match, which may lead to overestimating the system performance.
%
Some systems bypass this by comparing the sets and accepting a dialogue as matching if the sets are intersecting, or if the offered set is a non-empty subset of the ground-truth set. However, these differences result in large variances in the rates (see Section~\ref{sec:experiments}).

Another problem is related to the domain-oblivious delexicalization proposed by \citet{damd}. MultiWOZ responses contain slots from multiple domains at the same time very rarely, so it is sufficient to consider a single active domain for each turn. However, some works that adopt this new delexicalization use the ground-truth active domain during evaluation. Note that true domains have to be inferred from changes in ground-truth dialogue states and system actions.

\subsection{Output Diversity Metrics}
\label{sec:diversity}

The standard MultiWOZ metrics do not cover the diversity of the outputs, which can show the formulaic or repetitive nature of a system's responses \cite{holtzman_curious_2020}. While diversity is typically measured for non-task-oriented dialogue \cite{li_diversity-promoting_2016}, we argue that it can serve as an indicator of the naturalness of using a system over longer periods of time even in task-oriented dialogue such as MultiWOZ \cite{oraby_controlling_2018}.

\subsection{Dataset folds}
\label{sec:folds}

MultiWOZ authors split the data into train, validation, and test folds randomly. Following \citet{lampouras_imitation_2016}'s analysis of train-test overlap on other datasets, we inspected the goals of all 1000 test dialogues; 174 of them are also present in the train or validation folds. The test fold does not contain any unseen slot-value pairs, and has only 12 new domain-slot-value triplets. 
This means that the evaluation does not really check the generalization capabilities of the systems' state tracking,
and it theoretically allows the systems to memorize the whole database and bypass it during operation, which is a rather unrealistic assumption.

\section{Experiments}
\label{sec:experiments}

In this section, we work with outputs produced by all systems described in Section~\ref{sec:systems}. We: (1) unify their responses in terms of delexicalization styles, and then compare BLEU when different delexicalizations are applied, (2) evaluate Inform \& Success under identical conditions,\footnote{Note that we work with original authors' predictions, published pre-trained weights, or models trained from scratch, and thus we are not able to carry out a statistical analysis for the reported numbers.} (3) evaluate diversity and discuss similarity of the responses.

\subsection{Setup}
\label{ssec:setup}

We report BLEU scores for six different delexicalized references (see Table~\ref{tab:delex}). Five of them are styles used in HDSA, DAMD, AuGPT, UniConv, and LAVA. The sixth is delexicalization obtained from the MultiWOZ 2.2 span annotations. To make the BLEU-based comparison as fair as possible, we normalized the raw models' outputs. First, we remove start-of-sequence tokens, all “-s” and “-ly” strings and all “s” or “es” attached to a slot placeholder. Subsequently, we lowercase the utterances, identify slots names and map them to a unified slot name ontology. The ontology contains only 18 slot names (the original domain-aware delexicalization uses around 40 slot names). It is possible to map all the slot names used in the 6 different delexicalization styles onto it. To make a single mapping possible, the result is not lossless and reduces the finer level of detail provided by some systems. For example, slots named \emph{departure}, \emph{destination}, and \emph{taxi\_destination} are all replaced with the \emph{PLACE} placeholder.
Finally, we pass the utterances through Moses tokenizer and detokenizer\footnote{See https://github.com/alvations/sacremoses} \citep{koehn07}. To calculate BLEU, we use the SacreBLEU 
package\footnote{See https://github.com/mjpost/sacrebleu} \citep{post18}, which provides an implementation compatible with the original and is now a de-facto standard in MT (cf.~Section~\ref{sec:related}).

Inform \& Success rates depend on the database. Our database uses fuzzy matching for the different surface forms (see Section~\ref{sec:database}) using the FuzzyWuzzy package\footnote{See https://github.com/seatgeek/fuzzywuzzy} with a similarity threshold of 90\%. We use several rules to transform time strings, venue names, food types, and venue types to canonical forms matching the entries in the database (e.g., “ten o'clock p.m.” is replaced with “22:00”). 

Our implementation of the Inform \& Success rates follows the definition in Section~\ref{sec:mwoz-metrics}. The list of offered database entries, i.e. those complying to the current dialogue state, is updated only if a venue name or a train ID is mentioned (cf.~Table \ref{tab:setups}). Following HDSA, we accept a dialogue as matching if the set of offered entries is a non-empty subset of the set of entries matching the particular dialogue goal. Active domains of turns are taken from the original slot names if possible. If slot placeholders do not include the domain name, we either use model predictions if available, or estimate the domain from changes of state predictions in subsequent turns.

\begin{table*}[hbt]
    \centering\small
    \setlength{\tabcolsep}{2.6pt}
    \renewcommand{\arraystretch}{0.9}
    \begin{tabular}{l|ccccccc|cccccc}
      \toprule
      \multirow{3}{*}{Delexical.} & \multicolumn{7}{c|}{\multirow{2}{*}{End-to-end models}} 
             & \multicolumn{6}{c}{\multirow{2}{*}{Policy optimization models}}  \\
             & \multicolumn{7}{c|}{} 
             & \multicolumn{6}{c}{} \\
             & DAMD & MinTL & UBAR & SOLOIST & AuGPT & LABES* & DoTS & MarCo & HDSA & HDNO & SFN & UniConv & LAVA \\
      
      \midrule
      \textbf{MWZ 2.2} &  16.4 & 19.4 & 17.6 & 13.6 & 16.8 & \textbf{18.9} & \textbf{16.8} & 17.3 & 20.7 & 17.8 & 14.1 & 18.1 & 10.8 \\
      HDSA             &  15.5 & 18.6 & 16.3 & \textbf{15.1} & 15.5 & 17.1 & 15.7 & \textbf{19.0} & \textbf{22.5} & \textbf{19.4} & \textbf{15.6} & 17.9 & 11.4 \\
      DAMD             &  \textbf{16.9} & \textbf{20.0} & \textbf{17.9} & 14.1 & 16.5 & 18.7 & 16.7 & 17.8 & 21.4 & 18.3 & 14.6 & 18.3 & 11.0 \\
      AuGPT            &  15.8 & 18.6 & 16.7 & 13.2 & \textbf{17.0} & 17.9 & 16.6 & 17.1 & 20.4 & 17.7 & 13.5 & 18.0 & 10.5 \\
      UniConv          &  15.1 & 18.2 & 15.9 & 13.7 & 15.5 & 16.9 & 15.5 & 17.6 & 20.6 & 18.1 & 14.1 & \textbf{18.8} & 10.9 \\
      LAVA             &  15.4 & 18.6 & 16.3 & 15.1 & 15.5 & 17.1 & 15.7 & 19.0 & 22.5 & 19.4 & 15.6 & 17.9 & \textbf{11.4} \\
      \midrule
     \textbf{Reported} &  16.6 & 19.1 & 17.0 & 16.5 & 17.2 & 18.1 & 15.9 & 19.5 & 23.6 & 19.0 & 16.3 & 19.8 & 12.0 \\
     
      \bottomrule
    \end{tabular}
    \caption{Comparison of BLEU scores. The first column denotes the delexicalization style used for creating references. The highest score is highlighted for each system separately. The last row shows BLEU scores reported by authors. “*” denotes that scores for this system are computed on a subset of 91.66\% test utterances. }
    \label{tab:bleu}
\end{table*}

\begin{table*}[ht]
    \centering\small
    \setlength{\tabcolsep}{2.1pt}
    \renewcommand{\arraystretch}{1.0}
    \begin{tabular}{l|ccccccc|cccccc}
      \toprule
      \multicolumn{1}{c|}{\multirow{1}{*}{Metric}} & \multicolumn{7}{c|}{\multirow{2}{*}{End-to-end models}} &
      \multicolumn{6}{c}{\multirow{2}{*}{Policy optimization models}}  \\
      \multicolumn{1}{c|}{} & \multicolumn{7}{c|}{} & \multicolumn{6}{c}{} \\
      & DAMD & MinTL & UBAR & SOLOIST & AuGPT & LABES* & DoTS & MarCo & HDSA & HDNO & SFN & UniConv & LAVA \\
      \midrule
      \textbf{Inform}  &  57.9 & 73.7 & \textbf{83.4} & 82.3 & 76.6 & 68.5 & 80.4 & 94.5 & 87.9 & 93.3 & 93.4 & 66.7 & \textbf{95.9} \\
      Inform (rep.)    &  76.3 & 80.0 & 95.7 & 85.5 & 91.4 & 78.1 & 86.7 & 92.5 & 82.9 & 92.8 & 82.7 & 84.7 & 97.5 \\
      Inform (opt.)    &  73.7 & 79.3 & \textbf{88.6} & 86.1 & 78.1 & 75.8 & 84.4 & 96.9 & 91.6 & 97.7 & 96.7 & 67.5 & \textbf{97.5} \\
      \midrule
      \textbf{Success} &  47.6 & 65.4 & 70.3 & \textbf{72.4} & 60.5 & 58.1 & 68.7 & 87.2 & 79.4 & 83.4 & 82.3 & 58.7 & \textbf{93.5} \\
      Success (rep.)   &  60.4 & 72.7 & 81.8 & 72.9 & 72.9 & 67.1 & 74.2 & 77.8 & 68.9 & 83.0 & 72.1 & 76.3 & 94.8 \\
      Success (opt.)   &  63.0 & 71.1 & 75.0 & \textbf{76.2} & 62.4 & 65.5 & 74.4 & 89.9 & 83.2 & 90.2 & 87.0 & 60.1 & \textbf{95.9} \\
      \bottomrule
    \end{tabular}
    \caption{Comparison of Inform \& Success. “rep.” marks authors' reported results, “opt.” denotes results for the optimistic setting (see Section~\ref{ssec:setup}). “*” for LABES marks that scores were computed on 91.66\% of the test set.}
    \label{tab:rates}
\end{table*}

To better explain differences in the reported and our scores, we provide an \emph{optimistic} Inform \& Success following differences from the original implementation found in some systems, which can potentially overestimate results. In this setting, we: (1) use the intersection entry matching instead of subset matching, (2) ignore other search constraints if a name or ID is provided, (3) use ground-truth active domains.\footnote{We adopt the scripts for getting ground-truth active domains from DAMD's code base.} Note that (2) is more permissive with respect to the system's state tracking as the ground-truth context used during response prediction often contains ground-truth names or IDs. These are then used for the database search even if user constraints are not predicted correctly.

\begin{table*}[ht]
    \centering\small
    \setlength{\tabcolsep}{1.75pt}
    \renewcommand{\arraystretch}{1.2}
    \begin{tabular}{lr|rrrrrrr|rrrrrr}
      \toprule
      \multicolumn{1}{c}{\multirow{3}{*}{Measure}} & \multicolumn{1}{c|}{\multirow{3}{*}{Ref.}} & \multicolumn{7}{c|}{\multirow{2}{*}{End-to-end models}} & \multicolumn{6}{c}{\multirow{2}{*}{Policy optimization models}}  \\
      \multicolumn{1}{c}{} & \multicolumn{1}{c|}{} & \multicolumn{7}{c|}{} & \multicolumn{6}{c}{} \\
      
      & & DAMD & MinTL & UBAR & SOLO. & AuGPT & LAB.* & DoTS & MarCo & HDSA & HDNO & SFN & UC & LAVA \\
      \midrule
      Unique tokens       & 1407  & 212  & 297  & 478 & \textbf{615} & 608  & 374 & 411   & \textbf{319}  & 259  & 103  & 188  & 338  & 176  \\
      Unique trigrams     & 25212 & 1755 & 2525 & 5238 & \textbf{7923} & 5843 & 3228 & 5162 & \textbf{3002} & 2019 & 315  & 1218 & 2932 & 708  \\
      Entropy tokens      & 7.21  & 6.12 & 6.19 & 6.40 & 6.45 & \textbf{6.62} & 6.22 & 6.48 & \textbf{6.27} & 6.16 & 5.46 & 6.03 & 6.46 & 5.50 \\
      Con. ent. bigram    & 3.37  & 1.65 & 1.81 & 2.10 & \textbf{2.41} & 2.15 & 1.83 & 2.10 & \textbf{1.94} & 1.64 & 0.84 & 1.63 & 1.79 & 1.27 \\
      MSTTR-50            & 0.75  & 0.62 & 0.66 & 0.68 & 0.66 & \textbf{0.70} & 0.67 & 0.66 & 0.67 & 0.67 & 0.59 & 0.62 & \textbf{0.69} & 0.54 \\

      Avg. turn length & 14.07 & 14.27 & 14.78 & 13.54 & 18.45 & 12.90 & 14.20 & 14.66 & 16.01 & 14.42 & 14.96 & 14.93 & 14.17 & 13.28 \\
      \bottomrule
      
    \end{tabular}
    \caption{Comparison of lexical diversity measures. “Ref.” shows values for delexicalized MultiWOZ 2.2 references (see Section~\ref{sec:description}). Each system has its own column. “*” denotes that scores for this system are computed on a subset of 91.66\% test utterances. SOLO., LAB., UC stand for SOLOIST, LABES, and UniConv, respectively.}
    \label{tab:diversity}
\end{table*}

\subsection{Results}
\label{sec:results}

\paragraph{BLEU:} Table~\ref{tab:bleu} summarizes BLEU evaluation using different reference texts.
We notice that using a different delexicalization might substantially change the score (up to 2\% BLEU absolute). 
Most systems perform best on the references produced by their native delexicalization used for training. 
We can also see that different delexicalization styles result not only in different absolute values, but also in a different relative ordering of the systems.
This shows that having a single standard delexicalization (which should always be used for model evaluation and score comparison, and preferably also during model development) is very important for any fair comparison between the models.
Unlike in the case of end-to-end systems, the reported scores of the policy optimization models are higher then ours. 

\paragraph{Inform \& Success rate:}
Table~\ref{tab:rates} shows our and reported numbers for Inform \& Success. The corpus data, i.e. ground-truth responses and dialogue states, yield Inform 93.7\% and Success of 90.9\%. When evaluating in the optimistic setup, these numbers grow to 97.9\% and 96.6\%, respectively.

Our numbers differ from the reported scores of end-to-end models to a large degree, e.g.,
DAMD's reported performance is around 20\% higher for both rates. 
However, the optimistic setting results in much lower differences.
This shows that DAMD has problems with DST, which is hidden in the optimistic setup. 
The original UBAR numbers are very high because some ground-truth data were used during evaluation.
AuGPT reports higher rates caused by a different Inform rate computation, where the set of offered venues is obtained only at the end of the dialogue. 
Our scores are similar to the reported ones for SOLOIST and DoTS.
UniConv has the most different rates among the policy optimization models (ca.~17\% for both metrics). 
LAVA reports higher rates similar to ours in the optimistic setting, but the difference is small and may be caused by MultiWOZ version differences.
Our rates for SFN are much higher than the reported. 
MarCo's and HDSA's difference in rates can be accounted to our more flexible database. 


\subsection{Evaluating Diversity}
While the scores and rates differ between the evaluated systems, the generated utterances are similar and uniform (cf. Appendix~\ref{sec:appendixb}). To further understand differences between the systems, we analyzed the diversity of their responses (see Table~\ref{tab:diversity}). 

We compare the texts on several diversity measures, following \citet{van_miltenburg_measuring_2018} and \citet{dusek_evaluating_2020}: number of unique output tokens and trigrams, Shannon entropy and bigram conditional entropy, mean segmental type-token ratio (MSTTR-50),\footnote{MSTTR measures the average type-token ratio over the output text cut into segments of equal length (50 in our case). This reduces dependency on the overall text length, which is very strong in regular type-token ratio.} and average output length. We used the normalized texts with unified slot ontology (see Section~\ref{sec:results}) for the comparison. The ground-truth responses with MultiWOZ 2.2 delexicalization were used as reference. Even though the systems use different delexicalization schemes, we can draw some conclusions from the analysis. First, all the systems use rather small vocabularies. The number of used trigrams is orders of magnitude lower compared to human-produced texts. The bigram conditional entropy is also much lower for all systems. Models which employ reinforcement-learning, i.e. HDNO, SFN, and LAVA, produce the least diverse outputs. HDNO uses only 315 trigrams, which is around 1.2\% of the distinct trigrams seen in reference texts. On the other hand, AuGPT, UBAR, and DoTS seem to use a broader range of expressions. Extraordinarily diverse and long are the outputs of SOLOIST. However, they are still much more closer to other models then to the human reference.

\section{Conclusion}

The MultiWOZ benchmark is unique for its size and the inclusion of a complete database, making it possible to build end-to-end task-oriented dialogue systems. Because of its naturalness and thanks to multiple fixes and revisions of state annotations, it became very popular for dialogue state tracking. However, it still has limitations for context-to-response generation, partially because of lack of standardized preprocessing and postprocessing. Since standard, easy-to-use evaluation scripts are not available, researches are motivated to include their own modifications. This may appear unimportant, but as we showed in our analysis of 13 systems' outputs, it results in large differences in scores and makes any comparison or tracking of progress in this area problematic. 

We contribute to the solution of this problem by releasing 
evaluation scripts, which allow consistent evaluation of future work. We further include the evaluation of output diversity, which adds an important aspect missing from corpus-based MultiWOZ evaluation so far.

Future work should include a manual revision of MultiWOZ~2.2 
span annotation to reduce training noise and to enable fair evaluation on 
lexicalized outputs.
More important, however, is the use of human evaluation and evaluation of full dialogues in addition to corpus-based metrics \cite{liu16,takanobu_is_2020}, which is still not standard for  end-to-end dialogue systems (cf.~Section~\ref{sec:systems}). 

\section*{Acknowledgements}

We thank the reviewers for their kind feedback. 
This work was supported by the Charles University grants PRIMUS/19/SCI/10, GAUK 373921, and SVV 260~575.


\bibliographystyle{acl_natbib}
\bibliography{acl2021}

\appendix

\begin{table*}[t!]
    \centering\small
    \setlength{\tabcolsep}{4pt}
    \renewcommand{\arraystretch}{1.75}
    \begin{tabular}{p{0.5\linewidth}|p{0.145\linewidth}|c|c|c}
      \toprule
      \multicolumn{1}{l}{\textbf{Goal database entries (ID):}}   & \multicolumn{4}{p{0.4\linewidth}}{19212, 19185, 19197, 19219} \\
      \multicolumn{1}{l}{} & \multicolumn{4}{p{0.4\linewidth}}{(cheap and Chinese search constraints)} \\
      \multicolumn{1}{l}{\textbf{Goal requested information:}} & \multicolumn{4}{l}{ADDRESS, POST} \\
      \midrule
      \multirow{2}{*}{\textbf{Turn utterances}} & \multirow{2}{*}{\textbf{Dialogue state}} & \bf Offered & \bf Active & \bf Provided \\
       & & \bf Entries & \bf Domain & \bf Info \\
      
      \midrule
      \emph{User:} I'd like to find a resturant that serves cheap Chinese food! & \multirow{5}{2.25cm}{food: chinese, pricerange: cheap} & \multirow{5}{*}{no update} & \multirow{5}{*}{restaurant} & \multirow{5}{*}{$\O$} \\
      \emph{Ground-truth system:}  Charlie Chan is located in the centre and it is cheap! Would you like me to book it for you? & & & & \\
      \emph{System:} there are COUNT PRICE FOOD restaurant. do you have a preference of area? & & & & \\
      
      \midrule
      \emph{User:} Yes, go ahead and book it and please send me the address. & \multirow{5}{2.25cm}{food: chinese, pricerange: cheap, area: centre} & \multirow{5}{1cm}{19212, 19185, 19219} & \multirow{5}{*}{restaurant} & \multirow{5}{*}{ADDRESS} \\
      \emph{Ground-truth system:}  The address is Regent Street City Centre. What day and time would you like to book? How many people? & & & & \\
      \emph{System:} okay, i have booked NAME, which is located at ADDRESS. is there anything else i can do for you today? & & & & \\
      
      \midrule
      \emph{User:} I do not need to book it. What is the postcode and the area please? & \multirow{5}{2.25cm}{food: chinese, pricerange: cheap, area: centre} & \multirow{5}{1cm}{19212, 19185, 19219} & \multirow{5}{*}{restaurant} & \multirow{5}{1.5cm}{POST, ADDRESS} \\
      \emph{Ground-truth system:}  The postcode is cb21db & & & & \\
      \emph{System:} NAME is in the AREA and the postcode is POST. is there anything else i can help you with today? & & & & \\
      
      \midrule
      \emph{User:} Thanks very much! & \multirow{4}{2.25cm}{food: chinese, pricerange: cheap, area: centre} & \multirow{4}{*}{no update} & \multirow{4}{*}{-} & \multirow{4}{1.5cm}{POST, ADDRESS} \\
      \emph{Ground-truth system:}  You're welcome! Have a great day! Goodbye. & & & & \\
      \emph{System:} thank you for using our services. & & & & \\
      
      \bottomrule
      
    \end{tabular}
    \caption{Inform \& Success calculation on the predictions of the UBAR model (conversation ID: SNG0805).}
    \label{tab:walkthrough}
\end{table*}

\section{Inform \& Success Calculation Details}
\label{sec:appendixa}

Table~\ref{tab:walkthrough} walks through the process of Inform \& Success calculation.
Rows group conversation turns. The first column shows the last user utterance, the corresponding ground-truth system response and the delexicalized and normalized generated response. The second column shows the current dialogue state. The “Offered entities” column shows the changes of the set of matching venue or train IDs. Note that the set is updated only if the generated response contains the NAME or TRAINID placeholder. The “Active domain” column shows the currently active domain. The “Provided Info” column lists requestable slots mentioned until the given point.

This sample conversation is (1) matching, i.e. contributes positively to the Inform rate, because the set of offered entities or database entries at the end of the dialogue is a non-empty subset of the set of goal database entries, and also (2) successful because it is matching and all requested information defined by the dialogue goal, i.e., the restaurant address and post code, were provided. Note that the ground-truth context is used to generate system utterances during the evaluation, and thus the dialogue state might contain information from past ground-truth system utterances (such as the area slot in our example, which was never mentioned in the generated system or user utterances).

\section{Generated Texts Comparison}
\label{sec:appendixb}

Tables~\ref{tab:side-by-side1} and~\ref{tab:side-by-side2} give a comparison of delexicalized utterances generated by all examined systems for two different contexts. Slot placeholders are shown in capital letters. While there are some differences in the systems' behavior, the outputs often contain very similar phrasing.

\begin{table*}[t]
    \centering\small
    \setlength{\tabcolsep}{4pt}
    \renewcommand{\arraystretch}{1.4}
    \begin{tabular}{p{0.08\linewidth}|p{0.86\linewidth}}
      \toprule
      \multicolumn{1}{l}{\textbf{Context}} & User: Hello, I am doing research on plant life in England, and visiting cambridge as part of that, can you give me more info on the cambridge university botanic gardens. \\
      
      \multicolumn{1}{l}{\multirow{2}{*}{\textbf{System}}} & \multicolumn{1}{l}{\multirow{2}{*}{\textbf{Normalized utterance}}} \\
      \multicolumn{1}{l}{} & \\
      \midrule
      Corpus   & that sounds like an exciting trip! NAME is located in the AREA on ADDRESS. its admission fee is PRICE. \\
      \midrule
      DAMD       & the address is ADDRESS and the postcode is POST. is there anything else i can help you with? \\
      MinTL      & NAME is a TYPE in the AREA. it is located at ADDRESS. their phone number is PHONE. \\
      UBAR       & NAME is a TYPE located in the AREA at ADDRESS. it is PRICE to enter and is located on ADDRESS. is there anything else i can help you with? \\
      SOLOIST    & certainly! NAME is located at ADDRESS. their phone number is PHONE. is there anything else i can help you with? \\
      AuGPT      & NAME is a TYPE located in the AREA of town. the entrance fee is 4 pounds. the address is ADDRESS. \\
      LABES      & NAME is located in the AREA area of town \\
      DoTS       & NAME is a TYPE in the AREA. it's free to enter. do you need any other information? \\
      \midrule
      MarCO      & NAME is a nightclub in the AREA of town. the address is ADDRESS and the postcode is POST. \\
      HDSA       & it is a museum in the AREA of town. the entrance fee is free. would you like the address? \\
      HDNO       & NAME is a museum in the AREA. the address is ADDRESS. the phone number is PHONE. is there anything else i can help you with? \\
      SFN        & NAME is located at ADDRESS. the postcode is POST. the phone number is PHONE. can i help you with anything else? \\
      UniConv    & sure! it is a TYPE in the AREA part of town. would you like the address? \\
      LAVA       & the NAME is located at ADDRESS, postcode POST. the phone number is PHONE. \\
       \bottomrule
    \end{tabular}
    \caption{Side-by-side comparison of system outputs for the 1st system turn of the conversation PMUL3012.}
    \label{tab:side-by-side1}
\end{table*}
\begin{table*}[t]
    \centering\small
    \setlength{\tabcolsep}{4pt}
    \renewcommand{\arraystretch}{1.4}
    \begin{tabular}{p{0.08\linewidth}|p{0.86\linewidth}}
      \toprule
      \multicolumn{1}{l}{\textbf{Context}} & \begin{tabular}[t]{@{}p{\linewidth}@{}} User: I'm looking for a nice place to eat. \\ System: There are many fine places to eat. What type of cuisine would you prefer? \\ User: I don't care about the cuisine type, but I want it to be somewhere expensive and in the centre please.\end{tabular} \\
      
      \multicolumn{1}{l}{\multirow{2}{*}{\textbf{System}}} & \multicolumn{1}{l}{\multirow{2}{*}{\textbf{Normalized utterance}}} \\
      \multicolumn{1}{l}{} & \\
      \midrule
      Corpus   & there are COUNT PRICE eateries in the AREA. how about one of my favorites - NAME? they serve FOOD food and it's delicious! \\
      \midrule
      DAMD       & how about NAME? it is an FOOD restaurant that is in the AREA of town and is in the PRICE price range . \\
      MinTL      & there are COUNT restaurant that meet your criteria. do you have a preference for type of food? \\
      UBAR       & there are COUNT PRICE restaurant in the AREA. do you have a cuisine preference? \\
      SOLOIST    & how about NAME? it is an FOOD restaurant that is in the AREA of town and is in the PRICE price range . \\
      AuGPT      & there are 33 PRICE restaurants in the AREA. What type of food would you like? \\
      LABES      & i have COUNT restaurants that meet your criteria. \\
      DoTS       & i would recommend NAME. would you like to make a reservation? \\
      \midrule
      MarCO      & there are COUNT restaurant that meet your criteria. i recommend NAME. \\
      HDSA       & i recommend NAME. would you like a reservation? \\
      HDNO       & NAME is a FOOD restaurant in the AREA. would you like me to book a table for you? \\
      SFN        & there are COUNT PRICE restaurant in the AREA. what type of cuisine would you like? \\
      UniConv    & there are 33 PRICE TYPE in the AREA. what type of food are you looking for? \\
      LAVA       & NAME is located at ADDRESS, postcode POST. the phone number is PHONE. \\
       \bottomrule
    \end{tabular}
    \caption{Side-by-side comparison of system outputs for the 2nd system turn of the conversation PMUL2489.}
    \label{tab:side-by-side2}
\end{table*}

\end{document}